\documentclass[runningheads]{llncs}
\usepackage[T1]{fontenc}
\usepackage{graphicx,verbatim, multirow, booktabs}
\usepackage{subcaption} % Required for subfigures
\usepackage{float}
\usepackage{tcolorbox}
\usepackage{alphabeta}
\usepackage{fancyvrb}
\usepackage{fvextra}
\usepackage{colortbl}
\colorlet{tableheadcolor}{gray!75}
\colorlet{tablerowcolor}{gray!10}
\newcommand{\rowcol}{\rowcolor{tablerowcolor}} %

\usepackage{booktabs}  % Clean table format
\usepackage{multirow}  % Multirow support
\usepackage{graphicx}  % For text width scaling

\usepackage{orcidlink}

\usepackage[normalem]{ulem}
\useunder{\uline}{\ul}{}

\newcommand{\inlinesec}[1]{{\vspace{0.5em}\noindent\textbf{#1}}\hspace{0.5em}}

\begin{document}
\title{General Methods Make Great Domain-specific Foundation Models: A~Case-study on Fetal~Ultrasound}
\titlerunning{General Methods Make Great Domain-specific Foundation Models}
\newcommand{\repeatthanks}{\textsuperscript{\thefootnote}}

\author{Jakob Ambsdorf\inst{1,6}\thanks{Equal contribution.}\hspace{2pt}\orcidlink{0000-0003-2925-8809} \and
Asbjørn Munk\inst{1,6}\repeatthanks\hspace{2pt}\orcidlink{0009-0009-8170-079X} \and
Sebastian Llambias\inst{1,6}\hspace{2pt}\orcidlink{0009-0003-1898-3690} \and Anders Nymark Christensen\inst{2}\hspace{2pt}\orcidlink{0000-0002-3668-3128} \and Kamil Mikolaj\inst{2}\hspace{2pt}\orcidlink{0000-0002-1631-9329} \and Randall Balestriero\inst{5}\hspace{2pt}\orcidlink{0000-0002-5692-4187} \and Martin Tolsgaard\inst{1,3,4}\hspace{2pt}\orcidlink{0000-0001-9197-5564} \and Aasa Feragen\inst{2}\hspace{2pt}\orcidlink{0000-0002-9945-981X} \and Mads Nielsen\inst{1,6}\hspace{2pt}\orcidlink{0000-0003-1535-068X} }

\authorrunning{Ambsdorf et al.}
\institute{University of Copenhagen, Copenhagen, Denmark \and Technical University of Denmark, Kgs Lyngby, Denmark \and Copenhagen Academy of Medical Education and Simulation (CAMES), Rigshospitalet, Copenhagen, Denmark \and Copenhagen University Hospital Rigshospitalet, Copenhagen, Denmark \and Brown University, Providence, Rhode Island, USA \and Pioneer Centre for AI, Copenhagen, Denmark\\ 
\email{\{jaam,asmu\}@di.ku.dk}
}

\maketitle              % typeset the header of the contribution
\begin{abstract}

With access to large-scale, unlabeled medical datasets, researchers are confronted with two questions: Should they attempt to pretrain a custom foundation model on this medical data, or use transfer-learning from an existing generalist model? And, if a custom model is pretrained, are novel methods required? In this paper we explore these questions by conducting a case-study, in which we train a foundation model on a large regional fetal ultrasound dataset of 2M images. 
By selecting the well-established DINOv2 method for pretraining, we achieve state-of-the-art results on three fetal ultrasound datasets, covering data from different countries, classification, segmentation, and few-shot tasks. We compare against a series of models pretrained on natural images, ultrasound images, and supervised baselines. Our results demonstrate two key insights:
(i) Pretraining on custom data is worth it, even if smaller models are trained on less data, as scaling in natural image pretraining does not translate to ultrasound performance.
(ii) Well-tuned methods from computer vision are making it feasible to train custom foundation models for a given medical domain, requiring no hyperparameter tuning and little methodological adaptation. Given these findings, we argue that a bias towards methodological innovation should be avoided when developing domain specific foundation models under common computational resource constraints.

% Authors must provide keywords and are not allowed to remove this Keyword section.

\end{abstract}
\footnotetext[1]{Code available at: \url{https://github.com/jakobamb/UltraDINO}}

\section{Introduction}
\begin{figure}[t!]
    \centering
    \begin{subfigure}{0.48\linewidth}
        \centering
        \includegraphics[width=\linewidth]{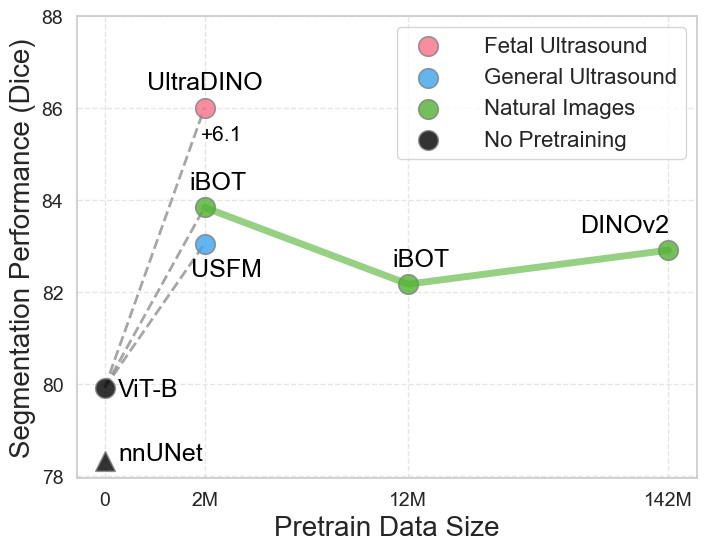}
        \caption{Dice score on JNU-IFM few-shot segmentation trained on four patients, with up to ten images per patient.}
        \label{fig:pretraining_vs_dsc}
    \end{subfigure}
    \hfill
    \begin{subfigure}{0.48\linewidth}
        \centering
        \includegraphics[width=\linewidth]{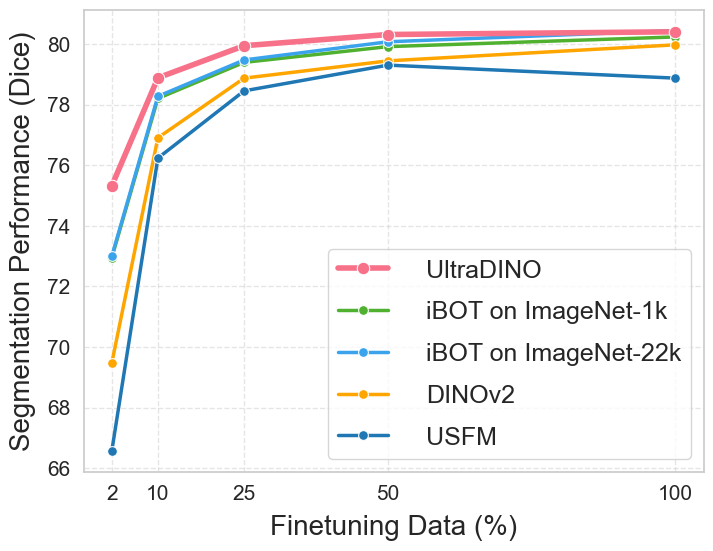}
        \caption{Few-shot performance vs. finetuning dataset size for pretrained models on segmenting fetal abdominal structures}
        \label{fig:fewshot_performance}
    \end{subfigure}
    \caption{Segmentation performance depending on model and pretraining dataset.}
    \label{fig:fass_comparison}
\end{figure}

Recently, the medical imaging community experienced a surge in interest in foundation models, which, compared to purely supervised baselines, provide a strong starting point for training models that achieve better performance, require fewer labels, and are useful as general-purpose feature extractors~\cite{bommasani2021opportunities,moor2023foundation,oquab2024dinov2,wald2024revisiting}. Foundation models are pretrained using self-supervised learning (SSL) on large-scale datasets without additional labels, and can be finetuned on downstream tasks with only a few annotated samples~\cite{krishnan2022self}.
Inspired by the success of these models in computer vision, various studies have investigated whether foundation models trained on natural images are useful for medical tasks as well~\cite{huix2024natural,song2024general}. However, the significant domain gap between natural images and the various medical imaging domains means that, while general visual features can be repurposed, the few-shot capability of these models remains limited, since semantics, such as patient anatomy, are not part of the data and are therefore not learned in pretraining~\cite{juodelyte2024source}. This initiated research on medical data in an attempt to learn more semantic features, modeling patient anatomy and properties of the imaging modality. These approaches use either a large mixture of data from various modalities and domains~\cite{wu2023towards}, where individual medical areas may not be well represented, or medium-scale, modality-specific 
datasets~\cite{jiang2024privacy,jiao2024usfm,kang2024urfm,wald2024revisiting}. A number of foundation models have since been proposed for medical ultrasound. USFM~\cite{jiao2024usfm} is trained on a highly unbalanced dataset of approximately 2 million ultrasound images and uses a Masked-Image-Modeling (MIM)~\cite{pathak2016context,he2022masked} approach with masking in both the spatial and the frequency domain. URFM~\cite{kang2024urfm} combines MIM on 1 million ultrasound images of different organs with knowledge distillation from a BiomedCLIP~\cite{zhang2023biomedclip} model. UltraFedFM~\cite{jiang2024privacy} is a concurrent work that uses a simple masked-autoencoder method in a multi-centre federated learning scenario on a diverse, 1 million image ultrasound dataset. 

Previous works on ultrasound foundation models used data from different anatomical regions, with fetal ultrasound not being well represented. At the same time, USFM and URFM introduced novel methodology, aimed at domain-specific improvements. In contrast, we investigate in this study how \textit{well-established} methods from computer vision can be used to train medical foundation models, when abundant data from a \textit{narrow domain} is available. We use the DINOv2 method to train UltraDINO, a foundation model for fetal ultrasound. UltraDINO is trained \textit{exclusively} on a regional fetal ultrasound dataset, containing 2 million high-quality, clinical images. We evaluate the effectiveness of UltraDINO on three fetal ultrasound datasets from Spain, Brazil, and China, with tasks spanning standard-plane classification, coarse and fine anatomical segmentation, and various few-shot settings. Through our experiments, we distill two key insights:

\begin{enumerate}
    \item It is worth the effort to pretrain on custom data, even if smaller models are trained on less data, as scaling in natural image pretraining does not translate to ultrasound.
    \item Well-tuned methods from computer vision are making it feasible to train custom foundation models for a given medical domain, requiring no hyperparameter tuning and little methodological adaptation.
\end{enumerate}

Given these findings, we argue that a bias towards methodological innovation on the pretraining method can be detrimental, as tuning novel pretraining methods needs extensive hyperparameter optimization, requiring large compute resources not commonly available for medical imaging groups operating in specialized domains. In contrast, this study demonstrates that directly applying well-established pretraining methods to a sufficiently large, domain-specific dataset reaches state-of-the-art results for fetal ultrasound and generalizes well across tasks and countries.

\section{Pretraining}
\begin{figure}[t]
    \centering
    \includegraphics[width=0.85\linewidth]{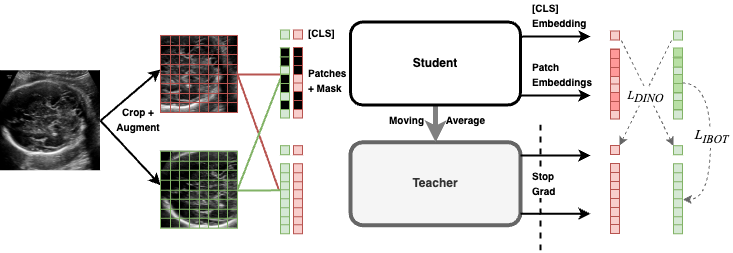}
    \caption{UltraDINO uses the self-distillation method used by DINOv2, image adapted from \cite{zhou2021ibot}}
    \label{fig:ibot}
\end{figure}

\subsubsection{Method.} UltraDINO uses the self-distillation~\cite{grill2020bootstrap} method of iBOT~\cite{zhou2021ibot} and DINO~\cite{caron2021emerging}, which was later refined by DINOv2 \cite{oquab2024dinov2}, as shown in Figure \ref{fig:ibot}. The method is based on the following idea: Given two image encoders, called student and teacher, and two views of the same image (with different crops, augmentations, or maskings), the student attempts to predict both global (\texttt{[CLS]}-token) and local (masked patch-tokens) embeddings of the teacher. To prevent collapse, regularizations are applied, and the teacher is constructed as an exponential moving average of the student's parameters~\cite{grill2020bootstrap}.

\inlinesec{Data.} Our pretraining dataset, denoted \textbf{FUS2M}, was collected the Danish national fetal ultrasound database, involving data from pregnant women who participated in prenatal ultrasound screening program in three out of five Danish regions between January 1st 2008 and June 30th 2018.  The screening program consists of two ultrasound examinations, captured in the first and second trimester. We retrieved a random sample of 2 million ultrasound images with approvals from The Danish Data Protection Agency (Protocol No. P-2019-310) and by The Danish Patient Safety Authority (Protocol No. 3-3031-2915/1) from regional servers, excluding around 10.000 images that were used in previous studies. The fetal ultrasound examinations were performed using GE Logiq 7 (GE Healthcare, Milwaukee, WI, USA) or Voluson E6, E8, and E10 machines (GE Healthcare, Zipf, Austria).

\inlinesec{Configuration.} Our model uses ViT-B and ViT-S vision transformer architectures~\cite{dosovitskiy2020image} with a patch size of 16, pretrained from scratch on FUS2M over 40 epochs, with a batch size of 512. All models were trained on two H100 80GB GPUs.

\section{Experiments}
We evaluate the pretrained models and non-pretrained baselines on one classification and two segmentation tasks. This includes evaluation of pretrained models using linear probing on classification, as well as full finetuning and comparison to non-pretrained baselines. We further evaluate the few-shot performance of models in different configurations. We report average results on the test set over three random seeds for full dataset experiments, and over five training splits in the few-shot settings. None of the evaluation datasets are included in FUS2M.

\subsection{Baselines}
We compare UltraDINO to the previously published general ultrasound foundation model USFM, as well as iBOT and DINOv2 which are trained on natural images. Since USFM did not publish code to reproduce their pretraining, we were unable to reproduce the method on FUS2M, and instead limit our evaluation to the published pretrained weights. USFM is finetuned using the recipes provided by the authors, since this yielded better results for the USFM model than using the configuration used for UltraDINO. For iBOT and DINOv2, we use official model checkpoints, which are trained on natural image datasets of different sizes, and on ViT-S and ViT-B models without pretraining. The iBOT models are trained on ImageNet-1K and ImageNet-22K consisting of 1.3M and 12.5M natural images respectively. DINOv2 is trained on LVD-142M consisting of 142M natural images. The DINOv2 checkpoints are distilled from a ViT-g model with 1.1B parameters. All models use patch size 16, except DINOv2 which is only available with patch size 14. Lastly, we evaluate non-pretrained performance of nnUNet~\cite{isensee2021nnu}, as well as other UNet variants outside of the nnUnet framework for segmentation performance. As the best performing supervised baseline, only nnUNet is also evaluated for few-shot.

\renewcommand{\arraystretch}{1.2}  % Slightly increased row height for readability

\begin{table}[t]
\centering
\caption{\textbf{Performance comparison of models across classification and segmentation tasks.} Results are averaged over three folds for full finetuning and five folds for few-shot learning. UltraDINO achieves superior few-shot learning and classification. JNU-IFM segmentation shows the largest performance gap. Linear probing suggests that UltraDINO embeddings are nearly linearly separable for standard planes.}
\label{tab:results}
\resizebox{\textwidth}{!}{%
\begin{tabular}{
    p{0.4cm}
    l
    p{2cm}
    *{3}{p{1.5cm}}
    p{1.5cm}
    p{1.5cm}
}
\toprule
 & \textbf{Method (Backbone)} & \textbf{PT Data} &
 \multicolumn{3}{c}{\textbf{Segmentation (Dice$\uparrow$)}} &
 \multicolumn{2}{c}{\textbf{Classification (F1$\uparrow$)}} \\
\cmidrule(lr){4-6}\cmidrule(lr){7-8}
 &  &  &
 \shortstack{\textbf{JNU}\\($p{=}4$)} &
 \shortstack{\textbf{FASS}\\($n{=}20$)} &
 \shortstack{\textbf{FASS}\\($n{=}942$)} &
 Fine-tune & LP \\ \midrule

% ---------- PRETRAINED -------------------------------------------------
\rowcol\multirow{8}{*}{\rotatebox{90}{\footnotesize Pretrained}} &
   UltraDINO (\scriptsize{ViT-B/16}) & 2 M FUS &
\textbf{86.01} & \textbf{74.24} & \textbf{79.51} & 94.54 & \textbf{95.06} \\

\rowcol &  UltraDINO (\scriptsize{ViT-S/16}) & 2 M FUS &
85.58 & 73.29 & 79.14 & \textbf{94.74} & 94.17 \\

& USFM (\scriptsize{ViT-B/16}) & 2 M US &
83.05 & 66.48 & 78.86 & 93.94 & 42.78 \\

& DINOv2 (\scriptsize{ViT-B/14 Dist.}) & 142 M Nat. &
83.55 & 69.11 & 79.26 & 94.00 & 76.61 \\

& DINOv2 (\scriptsize{ViT-S/14 Dist.}) & 142 M Nat. &
82.84 & 68.63 & 78.62 & 93.59 & 74.95 \\

& iBOT (\scriptsize{ViT-B/16}) & 12.5 M Nat. &
82.17 & 72.01 & 79.24 & 94.16 & 83.01 \\

& iBOT (\scriptsize{ViT-B/16}) & 1 M Nat. &
83.84 & 71.94 & 78.91 & 94.02 & 85.24 \\

& iBOT (\scriptsize{ViT-S/16}) & 1 M Nat. &
81.94 & 72.23 & 79.09 & 94.01 & 83.98 \\ \midrule

% ---------- FROM SCRATCH ----------------------------------------------
\multirow{6}{*}{\rotatebox{90}{\footnotesize From Scratch}} &
nnUNet (\scriptsize{UNet}) & – &
78.33 & 64.93 & 76.55 & – & – \\

& ViT-S/16 & – &
79.59 & 62.54 & 77.95 & 85.08 & – \\

& ViT-B/16 & – &
79.92 & 63.28 & 78.53 & 84.79 & – \\

& MultiResUNet\cite{ibtehaz2020multiresunet} & – &
– & – & 71.19 & – & – \\

& UNet\cite{ronneberger2015u} & – &
– & – & 72.65 & – & – \\

& UNetR\cite{hatamizadeh2022unetr} & – &
– & – & 69.96 & – & – \\ \bottomrule
\end{tabular}}
\end{table}

\subsection{Segmentation}
Segmentation models are trained by attaching a segmentation decoder head to the model. We follow \cite{jiao2024usfm,zhou2021ibot,oquab2024dinov2} and use UperNet \cite{xiao2018uperhead} as decoder and fine-tune both encoder and decoder. For fair comparison, all models are trained on 224x224 resolution images. We use the dice similarity score to measure the segmentation performance. 

\subsubsection{Tasks.} To gauge the segmentation performance of the foundation models, we use the FASS dataset and the JNU-IFM dataset. 

\underline{FASS} consists of 1,588 ultrasound images from 169 term pregnant women captured at a hospital in Santa Catarina, Brazil~\cite{souza_da_correggio_2023}. It includes expert-annotated segmentations of four anatomical structures: the fetal abdominal aorta artery, intrahepatic umbilical vein, stomach, and liver. We use three configurations, in which the same 20\% of the images are reserved for testing: (1) \textit{full dataset} in a 80/20 train/val split, (2) \textit{few-shot} with $20$ images for training (approximately 2\% of the dataset) and the remaining for validation, (3) \textit{label-efficiency} using 10\%, 25\% and 50\% of the dataset for training and the remaining for validation.  

\underline{JNU-IFM} contains 6,224 images from 78 ultrasound-videos, recorded from 51 patients at the Intelligent Fetal Monitoring Lab of Jinan University, China \cite{lu2022jnu}. It features a relatively simple segmentation task of deliniating the pubic symphysis and fetal head. We limit our experiments to a few-shot setting where we train on only the data from 4 patients, with a maximum of 10 images sampled per patient. Approximately half of the dataset is held out for testing, and validation is performed on all images not used for training. We filter out all images on which at least one anatomical structure is not visible.

\subsection{Classification}
Classification models are trained on the Fetal Planes dataset \cite{burgos2020evaluation} using both full end-to-end fine-tuning and linear probing, using the F1 score to measure classification performance.

\inlinesec{Task.} The Fetal Planes dataset is a standard plane classification dataset containing 12,400 images from 1792 patients and six classes, recorded at Hospital Clinic and Hospital Sant Joan de Deu, Barcelona, Spain~\cite{burgos2020evaluation}. Standard planes are specific views of the fetal anatomy as defined by clinical criteria. We test on the provided split of 5272 images and divide the remaining training data into an 80/20 split of train and validation.

\inlinesec{Linear Probing.} We evaluated our pretrained model using linear probing by freezing its feature extractor and training a simple linear classifier on top of the learned representations. This allowed us to assess the quality of the pretrained features without fine-tuning the entire model. 
We follow the evaluation approach of~\cite{oquab2024dinov2} by training a series of linear classifiers with different configurations on the train split while keeping the backbone fixed, and evaluate the best linear classifier according to validaiton accuracy on the test set. 

\begin{figure}[bt]
    \centering
    \includegraphics[width=0.55\linewidth]{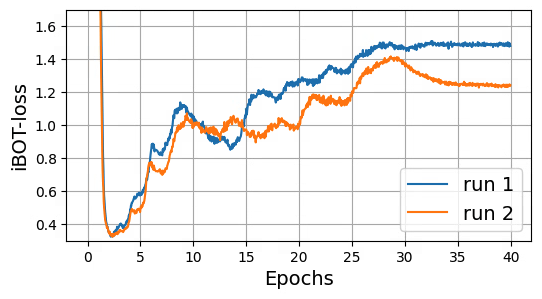}
    \caption{iBOT-loss curves for two runs with the same configurations \textit{and same seed} using a ViT-S on FUS2M are diverging considerably when using default non-deterministic algorithms.}
    \label{fig:loss}
\end{figure}

\section{Results and Discussion}

Table \ref{tab:results} provides an overview of the results. In general, UltraDINO models, both on small and base-size ViTs outperform all baselines in few-shot and classification tasks. On the segmentation tasks, the largest separation between UltraDINO and other models can be observed on the few-shot version of the JNU-IFM dataset. This is noteworthy, as the pubic symphysis/fetal head are, while being obstetric images, not part of the fetal screening protocol and therefore out-of-domain for the models trained on the FUS2M dataset. On FASS \textit{full}, all pretrained models reach similar performance, which could indicate other limiting factors, such as label noise and decoder precision. The \textit{few-shot} configuration is more informative for evaluating the embedding quality. Here, USFM is outperformed by all iBOT and DINOv2 baselines, while UltraDINO leads by 1-2 Dice points, depending on model size. The \textit{label efficiency} over different fractions of the FASS training dataset is represented in Figure~\ref{fig:fewshot_performance} and shows a similar trend, where UltraDINO reaches close to full performance on 25\% of the training data. Interestingly, larger pretraining datasets do not improve the performance of the natural image foundation models, as can be observed in Figure~\ref{fig:pretraining_vs_dsc}.

\begin{figure}[t]
    \centering
    \includegraphics[width=0.95\linewidth]{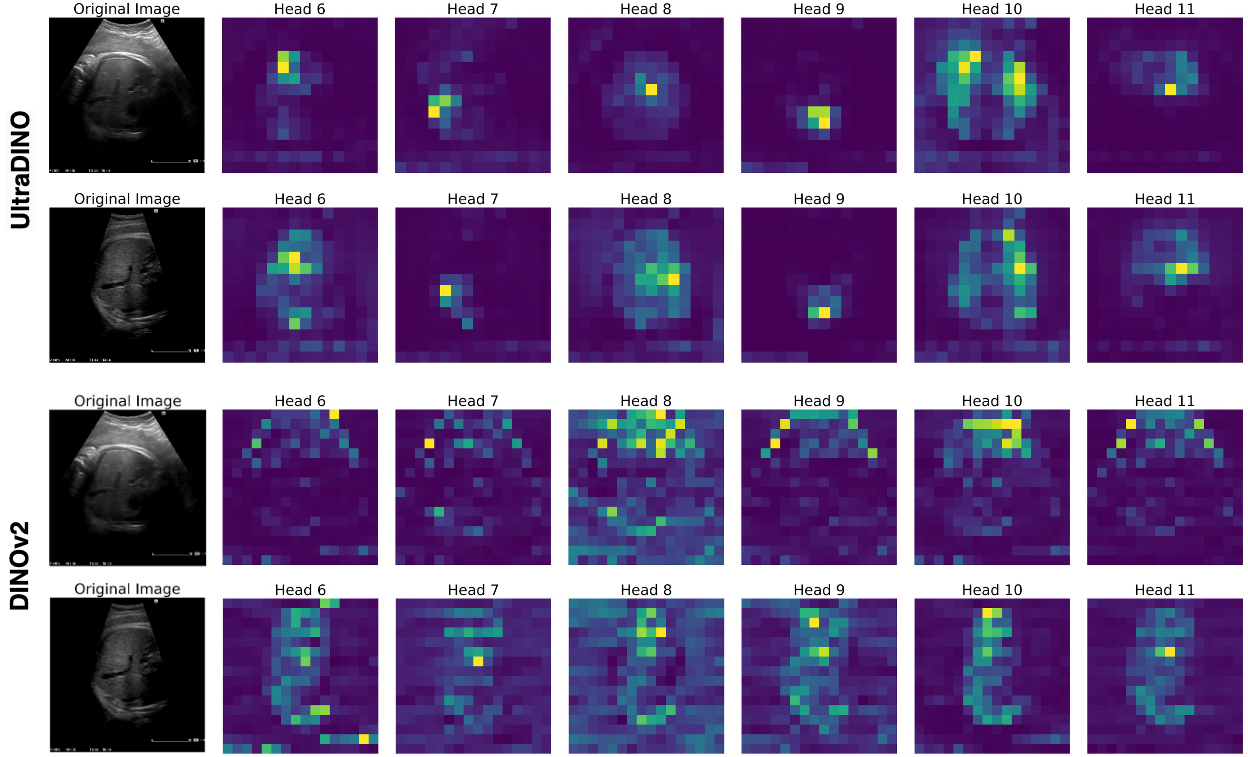}
    \caption{Attention maps of DINOv2 and UltraDINO ViT-B models: UltraDINO attention heads display semantic specialization to anatomical structures across images.}
    \label{fig:attention}
\end{figure}

After finetuning on the FetalPlanes dataset, the ViT-S of UltraDINO performs slightly better than the ViT-B, with all baselines trailing by about $0.5\%$ on F1. Linear probing reveals that UltraDINO embeddings are almost entirely linearly separable in the dataset classes. Surprisingly, all iBOT models perform very well and better than the DINOv2 models. USFM's performance on linear probing is poor, as it is common for masked auto-encoder-based models~\cite{he2022masked}.

Overall, any pretraining method is beneficial and significantly improves upon the non-pretrained baselines. USFM is outperformed by the natural image foundation models in the majority of cases, while UltraDINO models lead by a margin on few-shot and linear probing experiments. The results demonstrate that a well-tuned natural image foundation model outperforms domain-specific foundation models if the pretraining dataset is insufficient for the target domain and non-state-of-the-art pretraining methods are used, while the application of leading pretraining methods to high-quality pretraining datasets consistently improve upon natural image baselines. Notably, the pretraining dataset is not regionally overlapping with the finetuning tasks, indicating strong generalization capability despite this domain-shift.

\inlinesec{Tuning DINOv2 for fetal ultrasound.} During this study, we attempted to tune the hyperparameters of DINOv2 to match the ultrasound domain. We conducted experiments on ViT-S models with an iteration time of about 8 hours (at 318.39 GFLOPs per forward pass), but found that reaching confident conclusions on shorter training runs is inhibited by the high stochasticity intrinsic to the pretraining method. We observe that through large differences in the iBOT loss for the same model configuration, even when using the same random seed when using PyTorch's standard and significantly faster non-deterministic algorithms (see Figure~\ref{fig:loss}), adding another layer of complexity and computational cost for experimentation on the pretraining method. 

\inlinesec{Qualitative evaluation of attention maps.} Figure~\ref{fig:attention} shows the attention maps of UltraDINO and DINOv2 ViT-B models on the FASS validation set. We observe that while DINOv2 models capture high contrast edges, there is little to no specialization across attention heads. In contrast, UltraDINO attention maps are semantic and tend to relate to the same anatomical structures across images.

\inlinesec{Limitations.}
Our study has several limitations. Firstly, while we compare UltraDINO to USFM, DINOv2, and iBOT, we do not include other self-supervised learning (SSL) methods beyond self-distillation pretrained on natural or medical images. Secondly, the lack of available training code for USFM prevented us from reproducing its method on our dataset, limiting our analysis to the pretrained USFM weights provided by the authors. Further, while we evaluate on publicly available datasets from different countries, access to more diverse and standardized datasets would enable a more comprehensive assessment of generalization, especially for low-resource settings.

\section{Conclusion}
Through the case study of UltraDINO, we demonstrated using established methods that pretraining on a large-scale, high-quality dataset from a single medical domain reaches state-of-the-art performance and consistently outperforms generalist models trained on natural images, even when trained on substantially larger datasets.  
Our results emphasize two key findings: (i) Domain-specific pretraining is highly beneficial, and even smaller models trained on less data outperform larger models pretrained on natural images. (ii) The choice of pretraining method matters: By choosing a well-established self-supervised learning method, it is possible to achieve state-of-the-art performance across multiple fetal ultrasound tasks with no additional hyperparameter tuning. 
Based on these findings, and despite the computational cost and the stochastic variance of pretraining, 
% we argue that a focus on methodological innovation, rather than optimizing the dataset quality, can be counterproductive, as novel approaches often require extensive hyperparameter searches and large compute budgets. Therefore, 
we recommend training custom medical foundation models if relevant high-quality training data is available. 
However, instead of focusing on novel methodological development, well-tuned and established methods from computer vision are highly effective and should form the basis for any further experimentation.

    % The following acknowledgement and disclaimer sections should be removed for the double-blind review process.  
    % If and when your paper is accepted, reinsert the acknowledgement and the disclaimer clause in your final camera-ready version.

\begin{credits}
\subsubsection{\ackname} This work was supported by the Pioneer Centre for AI (DNRF grant nr. P1), the DIREC project EXPLAIN-ME (9142-00001B), the Novo Nordisk Foundation through the Center for Basic Machine Learning Research in Life Science (NNF20OC0062606) and Deep Fetal Development (NNF23OC0083562), the Danish Data Science Academy (funded by the Novo Nordisk Foundation
NNF21SA0069429 and Villum Fonden 40516), and SONAI, an AI signature project from the Danish Agency for Digital Government.

\subsubsection{\discintname}
ANC, MT, AF, and MN hold shares in Prenaital ApS.
% It is now necessary to declare any competing interests or to specifically
% state that the authors have no competing interests. Please place the
% statement with a bold run-in heading in small font size beneath the
% (optional) acknowledgments\footnote{If EquinOCS, our proceedings submission
% system, is used, then the disclaimer can be provided directly in the system.},
% for example: The authors have no competing interests to declare that are
% relevant to the content of this article. Or: Author A has received research
% grants from Company W. Author B has received a speaker honorarium from
% Company X and owns stock in Company Y. Author C is a member of committee Z.

\end{credits}

\bibliographystyle{splncs04}
\bibliography{ref}

\end{document}